\documentclass[submission,copyright,creativecommons]{eptcs}
\providecommand{\event}{Post Proceedings of ADG'23} 
\usepackage[T1]{fontenc}
\usepackage{xurl,amsmath}
\usepackage{cite}
\usepackage{graphicx}
\usepackage{amsfonts}
\usepackage{color}
\usepackage{booktabs}
\usepackage{amssymb}
\usepackage{doi}
\usepackage{hyperref}

\begin{document}
\title{The Locus Story of a Rocking Camel \\ in a Medical Center in the City of Freistadt}
\def\titlerunning{The Locus Story of a Rocking Camel in a Medical Center in the City of Freistadt}
\def\authorrunning{A.~Käferböck \& Z.~Kovács}
\author{
Anna Käferböck
\institute{The Private University College of Education of the Diocese of Linz, Austria}
\email{anna.kaeferboeck@gmx.at}
\and
Zolt\'an Kov\'acs 
\institute{The Private University College of Education of the Diocese of Linz, Austria}
\email{zoltan.kovacs@ph-linz.at}
}
\maketitle
\begin{abstract}
We give an example of automated geometry reasoning for an imaginary classroom project by using the free software package \textit{GeoGebra Discovery}. The project is motivated by a publicly available toy, a rocking camel, installed at a medical center in Upper Austria. We explain how the process of a false conjecture, experimenting, modeling, a precise mathematical setup, and then a proof by automated reasoning could help extend mathematical knowledge at secondary school level and above.
\end{abstract}

\section{Introduction}

Automated reasoning in geometry is available in various software tools for several years, mostly in prover packages. In this paper we pay our attention to a non-trivial presence of a geometry prover in the software tool GeoGebra Discovery \cite{EPTCS352.16,ACM2021} that aims at reaching secondary schools with its intuitive user interface.

Most importantly, we give a report on a STEM/STEAM project that was discussed in a group of prospective mathematics teachers at the Private University College of Education of the Diocese of Linz in Upper Austria during the winter semester 2022/23, in the frame of a course that focuses on exploiting technology in mathematics education (36 students in 2 working groups). This project consisted of several other experiments that were already communicated by the second author. The discussed activity, a detailed study of the movement of a rocking camel, is however, completely new. Also, some major improvements in the underlying software tool (implemented by the second author with a substantial help of the students’ feedback), makes it much easier to model similar project setups and conclude mathematical knowledge in an automated way.

\section{GeoGebra Discovery and its Automated Reasoning Tools}

GeoGebra Discovery is a fork of \textit{GeoGebra},\footnote{GeoGebra is an interactive geometry, algebra, statistics and calculus application, intended for learning and teaching mathematics and science from primary school to university level, available at \url{https://geogebra.org}.}\footnote{GeoGebra Discovery is available at \url{https://kovzol.github.io/geogebra-discovery}.} a de facto standard tool that supports mathematics education at various levels of learners. GeoGebra 5.0 and above come with a built-in automated reasoning subsystem. The supported commands: \textit{Prove}, \textit{ProveDetails}, \textit{LocusEquation} and \textit{Envelope} are further developed in GeoGebra Discovery by an addition of various other commands like \textit{Discover}, \textit {Compare} and \textit{RealQuantifierElimination}. Also, several improvements of the existing commands are included.

Most importantly, we focus on the symbolic support of the \textit{Dilate} command and tool that was added in GeoGebra Discovery version 2023Feb01.\footnote{See \url{https://github.com/kovzol/geogebra/releases/tag/v5.0.641.0-2023Feb01}.} It can be used to dilate an object from a point (which is the dilation center point), using a given factor, a rational number. This makes it easy to divide a line segment in a given ratio. Formerly, for such constructions the intercept theorem, or a consecutive use of midpoints, reflections or rotations had to be used. As a further result, the user can use the slider feature of GeoGebra (which is a numerical tool) and at the same time precise discovery and proofs (which are symbolic tools) can be automatically obtained now.

Our paper argues for the possible classroom use of GeoGebra Discovery on the one hand, and also for activities that combine real-life applications and automated geometry reasoning.

\section{A Rocking Camel}

\begin{figure}
\begin{center}
\includegraphics[width=0.6\linewidth]{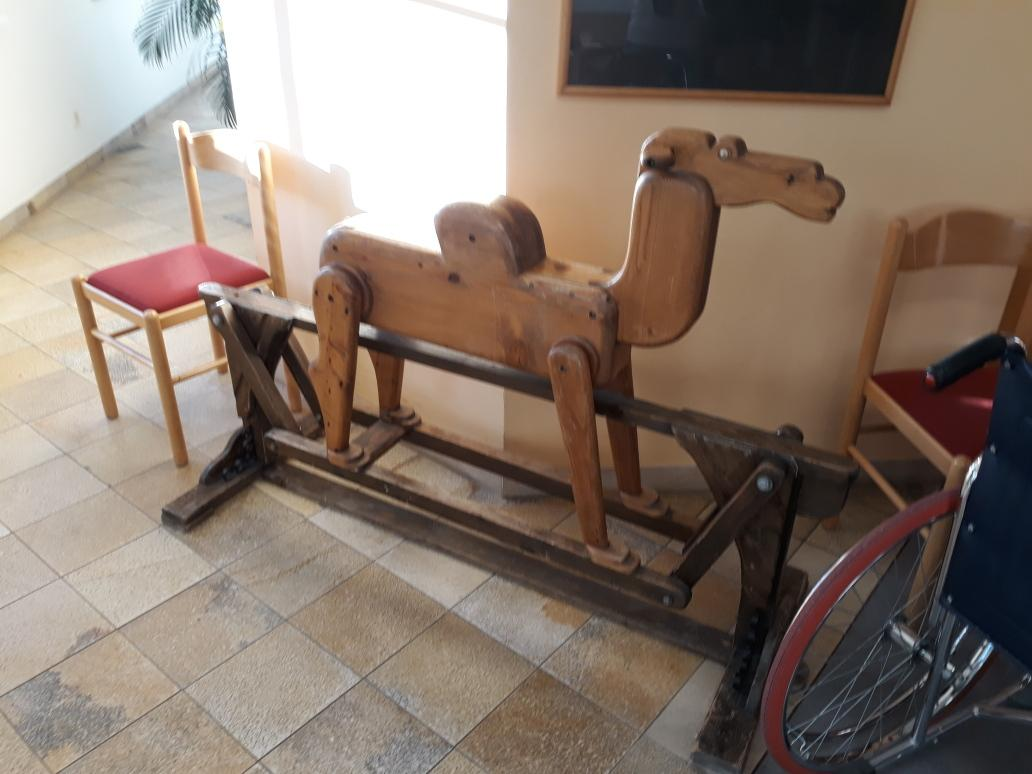}
\caption{The exhibited toy in the medical center of the city of Freistadt, Upper Austria.}
\label{f1}
\end{center}
\end{figure}

\begin{figure}
\begin{center}
\includegraphics[width=0.9\linewidth]{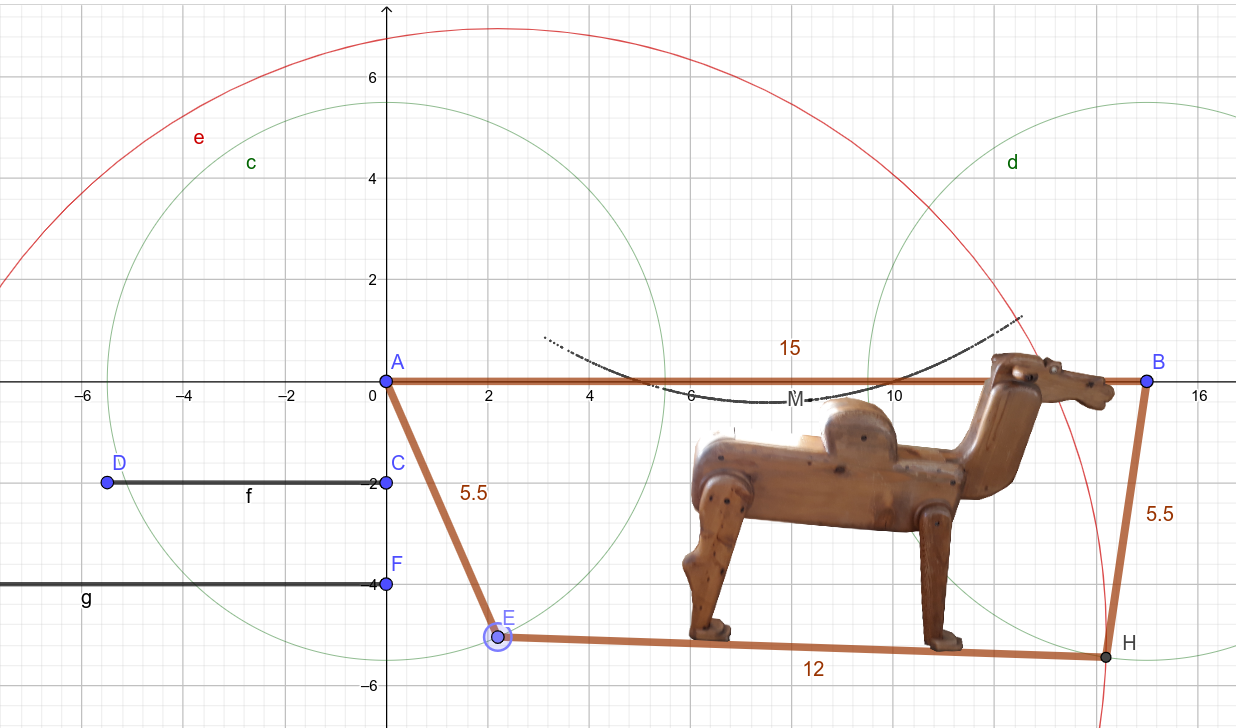}
\caption{A former version of a GeoGebra applet that suggests that the motion of the hump of the camel is a part of an ellipse.}
\label{f2}
\end{center}
\end{figure}

The toy shown in figure \ref{f1} is exhibited in a medical center in Freistadt, Upper Austria. It is installed for amusement purposes for children who are waiting for medical treatments. An obvious question is, from the mathematical point of view, to identify the movement of certain points of the camel.
Clearly, some points of the camel move on circular paths. For example, the suspension points, close to the legs, move along a circle. When asking for a general point of the camel, however, some non-trivial movements can show up. For example, the movement of the hump of the camel seems to move on an elliptical path, after the first experiments are performed by using a former version of the GeoGebra applet at \url{https://geogebra.org/m/b8mbjxcz} (Fig.~\ref{f2}).

Here we can also note that points $E$ and $H$ (they are the above-mentioned suspension points close to the legs of the camel) indeed move on a circular path. In fact, here the lengths of the quadrilateral $ABHE$ are given with the segments $f$ and $g$, and point $M$ was constructed by consecutive use of midpoints and a rotation by 90 degrees. At a later point of our paper we will use a better approach, based on the improvements on the Dilate command.

\subsection{The History of the Rocking Camel}

As a part of our project we researched after on how the camel got to the medical center. In fact, the camel was in the attic for many decades and no one knew what it was all about. It was probably passed down from generation to generation. Unfortunately, this much is known about it.

How many more toys are there in the attic that have a nice mathematical background but we have forgotten about them? How many forgotten mathematical books, writings and ideas are there that the modern age has put aside and not even superficially exploited their excitement?

We believe that our contribution will help dust off these forgotten gems and put them at the service of education today.

\section{A STEM Activity}

The abbreviation STEM stands for “science, technology, engineering and mathematics” \cite{Penprase21}. It is a popular approach to teach mathematics via real-life applications. Sometimes STEM is extended with an “A” (“arts”) and it becomes the abbreviation “STEAM”. Later we will learn how this engineering experiment can be extended to an artistic activity.

Now, the main question of the activity, raised to the prospective mathematics students, was to describe the movement of the hump of the camel by following the steps below:

\begin{enumerate}

\item Make an exact measurement of the toy and its parts. (As a first approach, this was prepared by a student by providing photos. Later, with another student, more exact data was collected in the medical center, by using measuring tapes, a camera and some graphical analysis with technology.)
\item Model the toy in GeoGebra and trace the movement of the hump. (The students already had an acceptable background of GeoGebra knowledge to make it possible to do experiments on their own.)
\item Make a conjecture. (Here most students conjectured that the movement was an ellipse.)
\item Show the locus of the trace points. (We will see later that the conjecture was wrong, because the trace shows a different curve, namely, something like a form “8”).
\item Make a second conjecture. (This was a very difficult question, since an 8-formed curve is not present in the curriculum, neither at secondary nor university level.)
\item Compute the mathematical equation of the locus. (This is easy by using the command or tool LocusEquation. Without this step, no satisfactory conjecture can be done.)
\item Check the conjectures. (This is possible by setting up an equation system by using pencil and paper, and then compute the locus curve by using technological means. For this problem, however, the students skipped this step. It was used to check a different problem, publicized in the LEGO 4094 set as the “moving monkey” \cite{oldenburg}.)
\item Generalize the problem with different inputs. (In general, we have a 4-bar linkage problem \cite{hunt} that leads to a sextic movement.)
\end{enumerate}

In the next subsections we give some details on the steps described above.

\subsection{Exact Measurements}

\begin{figure}
\begin{center}
\includegraphics[width=0.3\linewidth]{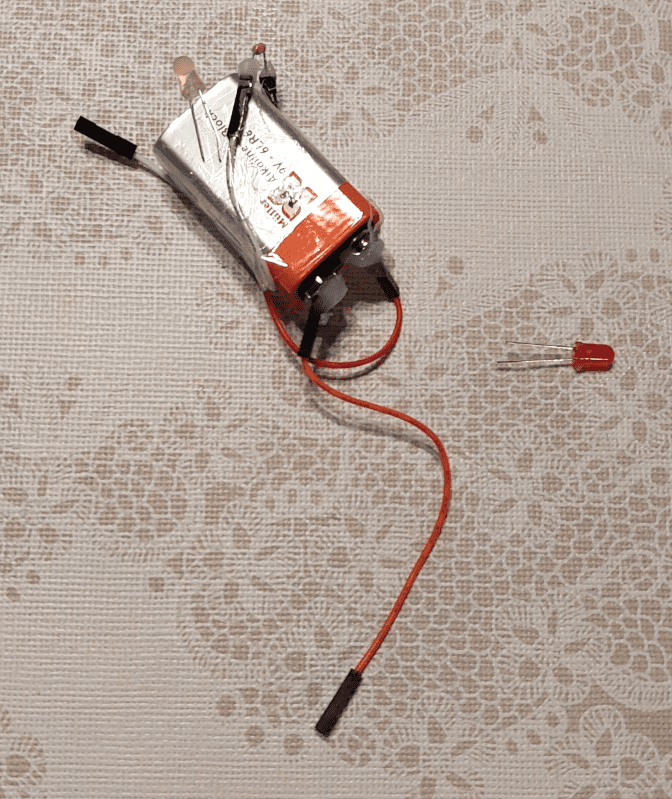}
\caption{A small lamp mounted on a battery with cables. It can be attached to a moving object by a glue tape.}
\label{f3}
\end{center}
\end{figure}

\begin{figure}
\begin{center}
\includegraphics[width=0.3\linewidth]{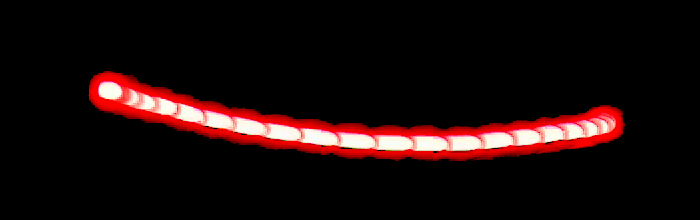}\\
\includegraphics[width=0.3\linewidth]{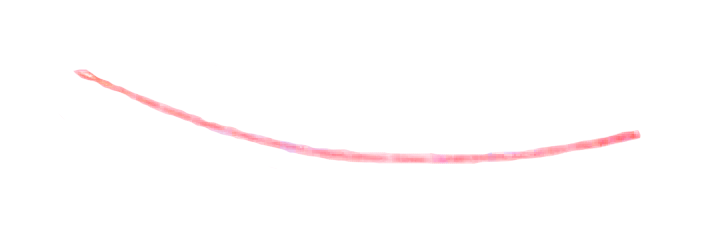}\\
\caption{Motion of the lamp, by using two steps of preprocessing.}
\label{f4}
\end{center}
\end{figure}

After measuring the distances among the most significant parts of the camel we mounted a small lamp with a battery on the camel (Fig.~\ref{f3}). Then we switched the light off and recorded the movement with a camera of a mobile phone.
When attaching the lamp to the top of the camel, we can get a motion like shown in figure \ref{f4}. These pictures were created after saving individual frames (25 images) with the \textit{VLC media player}\footnote{VLC media player is a free and open-source, portable, cross-platform media player software and streaming media server developed by the VideoLAN project, available at \url{https://www.videolan.org/vlc}.} and then opening them in \textit{GIMP}.\footnote{GIMP (GNU Image Manipulation Program) is a free and open-source raster graphics editor used for image manipulation (retouching) and image editing, free-form drawing, transcoding between different image file formats, and more specialized tasks, available at \url{https://www.gimp.org}.} Then the individual layers were edited with the “Exposure” function by changing the value of Black level to 0.1 (instead of 0.0). Furthermore the background of all layers was removed with the help of the function “Color by Alpha”, so that only the red light was left and the single layers did not cover each other anymore.

\subsection{Modeling in GeoGebra}

\begin{figure}
\begin{center}
\includegraphics[width=0.7\linewidth]{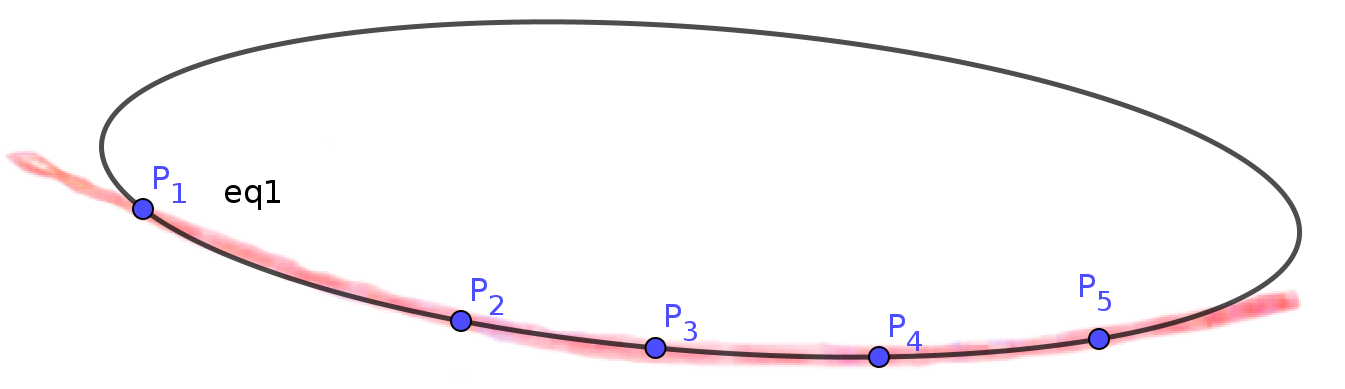}
\caption{An attempt to identify the motion as an ellipse.}
\label{f5}
\end{center}
\end{figure}

An option to continue with is to try to fit a curve on the output. This is well-supported in GeoGebra by the possibility to insert a transparent figure, making it as a background picture, and then create some free points by hand that approximately cover the curve. GeoGebra’s \textit{ImplicitCurve} command can find the best fitting implicit polynomial (see figure \ref{f5} or \url{https://geogebra.org/m/c93pegab} for an online applet): for a curve of degree $n$ one needs to enter $\frac{n\cdot(n+3)}2$ input points. That is, if we expect that the motion follows an ellipse (which is of degree 2), then 5 points are required.
During the university course, however, we followed a different path. In GeoGebra we constructed the drawing as in figure \ref{f2} by creating free points $A = (0,0)$, $B = (15,0)$, then creating a segment $CD$ with length $f = 5.5$, drawing a circle $c$ with center $A$ and radius $f$, and another circle $d$ with center $B$ and radius $f$. Then we attached point $E$ on $c$, and after this step we created another segment $FG$ with length $g = 12$. Next, we drew a third circle $e$ with center $E$ and radius $g$. One of the intersection points of $d$ and $e$ was designated to be point $H$. Then, as mentioned above, point $M$ was created with some further steps by halving and rotating some additional points.

We remark that this construction is a special case of a planar 4-bar linkage, which is well-known in the study of mechanisms, and has important applications like Watt’s steam engine or a pumpjack.

An exact GeoGebra model helped the students to make experiments with the linkage without visiting the medical center and making their own measurements.

\subsection{A First Conjecture}

The students had one week of working time to make a conjecture. Several learners made a false conjecture, however, because they had no idea that there could be a solution other than the ellipse. This also raises the general question of the pedagogical consequences of oversimplifying the mathematical modeling of world problems.

\subsection{A Numerical Locus}

\begin{figure}
\begin{center}
\includegraphics[width=0.7\linewidth]{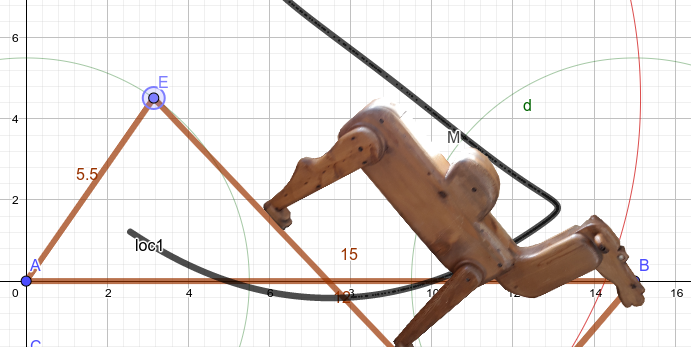}
\caption{Dragging point $E$ in an unrealistic position to disprove
that the searched motion is elliptical.}
\label{f6}
\end{center}
\end{figure}

Some students, however, continued dragging point $E$ to unrealistic positions and they obtained visual evidence that the searched curve is clearly not an ellipse (Fig.~\ref{f6}). This can also be checked in the above mentioned applet by enabling the “Locus” checkbox.

\subsection{A Second Conjecture}
At this point, a second conjecture could be made, but due to the lack of ideas, we more or less skipped this step. In fact, if you do not know the concept of higher degree curves, there is no chance to have a conjecture that the output is a polynomial curve.

\subsection{A Symbolic Locus}
\begin{figure}
\begin{center}
\includegraphics[width=0.5\linewidth]{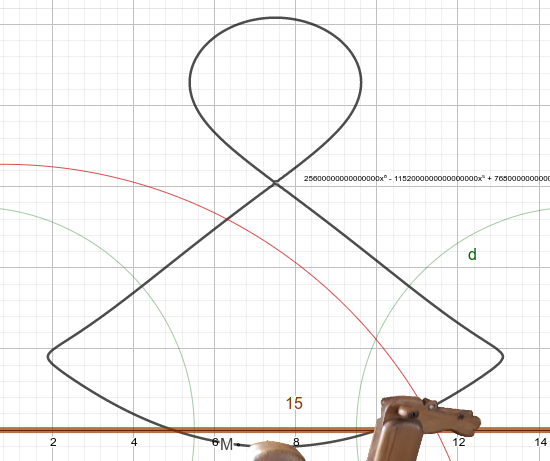}
\caption{Obtaining a symbolic locus equation.}
\label{f7}
\end{center}
\end{figure}

This step can be reproduced by enabling the “LocusEquation” checkbox in the above mentioned applet. We obtain, by using some computer algebra (which is not further explained in this step) a sextic polynomial equation,
$$256 \cdot  {10} ^ {14} {x} ^ {6} - 1152 \cdot  {10} ^ {17} {x} ^ {5} {y} ^ {2} - 768 \cdot {10} ^ {14} {x} ^ {4} {y} ^ {2} - 312 \cdot {10} ^ {15} {x} ^ {4} y + \ldots = 0$$ (Fig.~\ref{f7}). Here the students can only rely on the underlying computer algebra system, it is just a black box, but the coincidence of the numerical and symbolic loci can confirm, at least, partially, that the computations are hopefully correct.

\subsection{A Proof}

Now we need to prove that the obtained curve is indeed a sextic. To achieve this, we can set up an equation system with equations $a^2+b^2=5.5^2$ (here $E = (a,b)$), $(c-15)^2+d^2=5.5^2$ (here $H = (c,d)$), $(a-c)^2+(b-d)^2={12}^2$, and for obtaining the coordinates $M$ we might compute the coordinates of the midpoint $I$ of segment $EH$ and then rotate $E$ around $H$ by $-90$ degrees to get $E’$. Having $E’$, the midpoint $J$ of $IE’$ can help to create the midpoint $K$ of $JE’$, and midpoint $L$ of $KE’$. Finally, $M$ is the midpoint of $LM$. This process is, of course, quite complicated, but it shows how we can be arbitrarily close to any point of the camel, by using just simple geometric operations. Later, by using dilations, this will be easier.

Now, by using elimination from algebraic geometry we can obtain the locus equation by using Geo\-Gebra’s \textit{Eliminate} command. This is still a black box operation, but at least the students can have an idea what the exact input is, and the teacher can argue that by using the first three basic operations (addition, subtraction and multiplication), there is a finite algorithm \cite{gb-en} that indeed produces the result.

And this is actually a proof, in the deepest sense of the notion. Even if the atomic steps of the computation remain hidden, a reliable computer algebra system on reliable hardware will indeed compute the expected equation of the searched curve.

Let us highlight this fact even more. In classical geometry we are used to proofs that give arguments why the studied outputs are certain curves like lines, circles or maybe ellipses. The argumentation is sometimes purely synthetic, but sometimes analytic. Here we cannot really give a synthetic argumentation why a sextic curve appears. Only an analytic proof is applicable. But, because of the technical difficulty of the proof there is no way to check each step in a manual way. Therefore, a computer assisted proof is required, and as such, the automated way of elimination is satisfactory.

\subsection{Generalization}

With some feedback from the students it was possible to improve GeoGebra Discovery to support generalizing the problem setting in the following way: \textit{How does the output curve change when the lamp has a different position than the hump of the camel?}

\begin{figure}
\begin{center}
\includegraphics[width=0.9\linewidth]{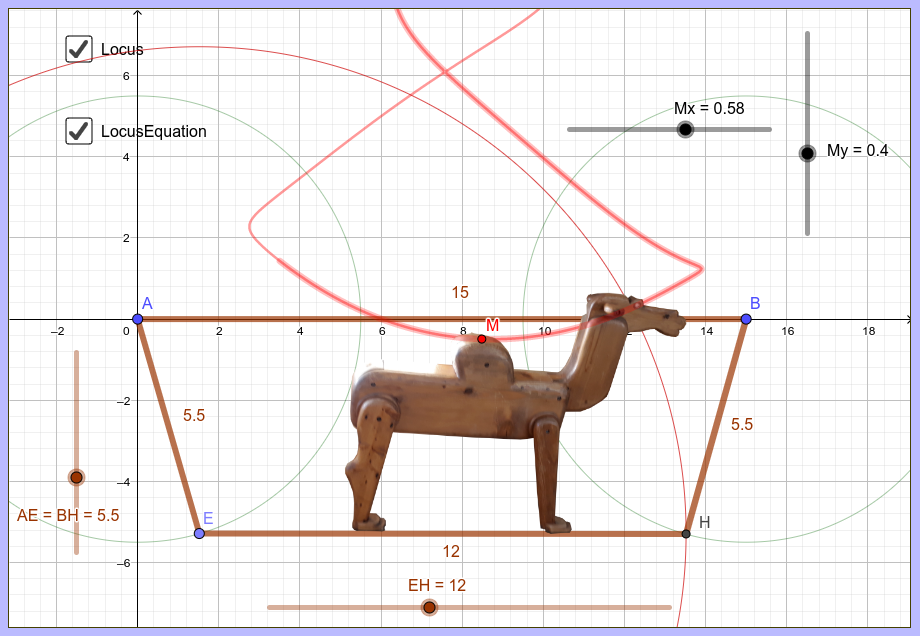}
\caption{Generalization with sliders via the Dilate command.}
\label{f8}
\end{center}
\end{figure}

\begin{figure}
\begin{center}
\includegraphics[width=1\linewidth]{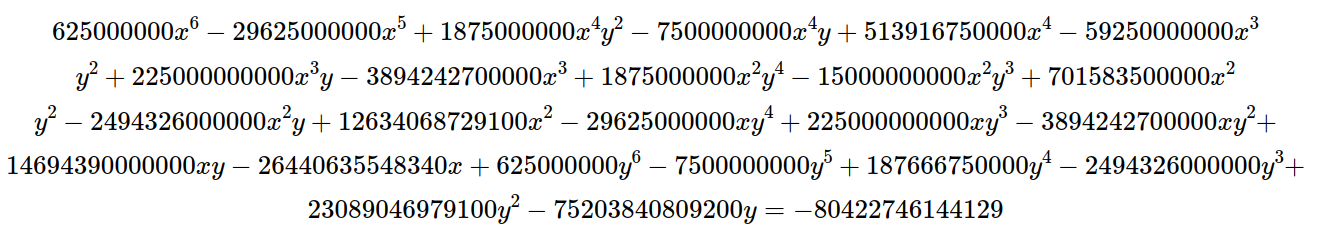}
\caption{A particular sextic equation, the corresponding curve is plotted in figure \ref{f8}.}
\label{f9}
\end{center}
\end{figure}

To achieve this, the Dilate command in GeoGebra required symbolic support. The applet at \url{https://matek.hu/zoltan/camel.php} (see figure \ref{f8}) allows the user to conveniently change the length of the bars $AE$ and $BH$ (they are still equally long) and the bar $EH$. By using dilation and sliders, the background computation requires less variables, because instead of 4 free variables just one needs to be used. This speeds up the computation substantially. To avoid the difficult way of defining $M$ we introduced two sliders $Mx$ and $My$ that help find the position of $M$ in an intuitive way. In addition, the user is notified immediately when the locus equation changes by using GeoGebra’s JavaScript API\footnote{Available at \url{https://wiki.geogebra.org/en/Reference:GeoGebra_Apps_API}.} (Fig.~\ref{f9}).
This applet was created by the use of the Dilate command. Dilation allows the user to create an arbitrary linear combination of two vectors. The coefficients of the linear combination can usually be rational numbers. Using one direct step to define ratios of certain quantities, instead of using the intercept theorem or utilizing midpoints, helps simplify the construction and avoid slow computation because of the high amount of variables. As well-known, elimination may be double exponentially slow in the number of variables in its worst case \cite{MayrMeyer82}. Therefore each optimization step may be crucial.

As a conclusion, the students can have a general conjecture after some further experiments, that 4-bar linkages usually yield sextic curves \cite{JSC-linkages}. Of course, such experiments are insufficient to get a general proof for all possible parameters. And, in fact, in some degenerate cases these results are actually not true, for example if the construction collapses into one point.

\section{Final Thoughts and Conclusion}

Automated geometric proofs may play an even more important role as before at secondary school level and above. The concept of analytic proofs (instead of synthetic ones) can already be familiar with algebraization of the geometric setup. For example, the well-known theorem by Thales that highlights a connection between right triangles and their circumcircles, can be easily translated into an algebraic setup and proven without difficulty. Indeed, let $A = (-1,0)$, $B = (1,0)$, $C = (x,y)$, and assume that $x^2+y^2=1$, that is, $C$ lies on a circle whose diameter is segment $AB$. Now, checking if $AC$ is perpendicular to $BC$ means exactly that $(x-(-1))\cdot(x-1)+(y-0)\cdot(y-0)=0$, and this is equivalent with our assumption on the sum of squares. That is, after making sure that the algebraization is performed correctly and generally enough, some algebraic manipulation will give the required argumentation.

Such an easy derivation is, unfortunately, not always possible. But we can learn that it is possible to formulate also the converse of the statement, that is, to ask: What is the geometric locus of points $(x,y)$ such that $AC$ is perpendicular to $BC$, when $A$ and $B$ are fixed? And here we conclude that the searched equation is $x^2+y^2=1$, a \textit{quadratic} one, in particular, the equation of a circle. In general, however, we may obtain \textit{non-linear} and \textit{non-quadratic results} as well. In our example in this paper we obtained a sextic equation, with huge coefficients. And this can happen in many other situations. Real life examples (of study of mechanisms, or optics) are full of higher degree polynomial curves. Here we mention conchoids, cissoids, strophoids (of degree 3) or cardioids, deltoids or lemniscates (of degree 4), many of them already well-known by the ancient Greek mathematicians.

In such higher degree cases, a proof that a certain curve is the expected result is nothing else than a long elimination process. Even if the computations are hidden, we expect that each step of the derivation is performed correctly, and therefore the result is correct.

That is, STEM/STEAM education cannot avoid such proofs in the long term. But, luckily, the existing tools are already safe and rich enough to support the learners in both the exploration and verification.

\section*{Acknowledgments}

We are grateful to students Eva Erhart and Engelbert Zeintl for their help in many aspects of this paper.
Benedek Kovács kindly helped us in preparing the photo of the rocking camel for further work in Geo\-Gebra. He also helped in the preparation of the mounted small lamp. The second author was partially supported by a grant PID2020-113192GB-I00 (Mathematical Visualization: Foundations, Algorithms and Applications) from the Spanish MICINN.

\bibliographystyle{eptcs}
\bibliography{refs}

\end{document}